\documentclass[12pt]{article}

\usepackage[utf8]{inputenc}
\usepackage{amsmath, amssymb}
\usepackage{graphicx}
\usepackage{geometry}
\usepackage{setspace}
\usepackage{hyperref}
\usepackage{authblk} % For author and affiliation block
\usepackage{cite} % For numbered Vancouver-style citations
\usepackage{placeins}
\usepackage{subcaption} 
\usepackage{forest} % For hierarchical tree structures
\title{SerendibCoins: Exploring The Sri Lankan Coins Dataset}

\author[1]{NH Wanigasingha}
\author[1]{ES Sithpahan}
\author[1]{MKA Ariyaratne\thanks{Corresponding author: \texttt{mkanuradha@sjp.ac.lk}}}
\author[1]{PRS De Silva}
\affil[1]{Department of Computer Science, Faculty of Applied Sciences, University of Sri Jayewardenepura, Sri Lanka}

\date{}

\begin{document}
	
	\maketitle
	
	\begin{abstract}
		The recognition and classification of coins are essential in numerous financial and automated systems. This study introduces a comprehensive Sri Lankan coin image dataset and evaluates its impact on machine learning model accuracy for coin classification. We experiment with traditional machine learning classifiers K-Nearest Neighbors (KNN), Support Vector Machines (SVM), and Random Forest as well as a custom Convolutional Neural Network (CNN) to benchmark performance at different levels of classification. Our results show that SVM outperforms KNN and Random Forest in traditional classification approaches, while the CNN model achieves near-perfect classification accuracy with minimal misclassifications. The dataset demonstrates significant potential in enhancing automated coin recognition systems, offering a robust foundation for future research in regional currency classification and deep learning applications.
	\end{abstract}
	

\section{Introduction}
Currency recognition is a critical task in various domains, including banking, retail, vending machines, and assistive technologies for visually impaired individuals \cite{shah2015review, nasir2024pakistani, ahmad2024deep, rafi2024nstu}. With the increasing adoption of automated systems and digital solutions, the ability to accurately identify and classify currency notes and coins has become essential. However, the development of robust currency recognition systems heavily relies on the availability of comprehensive and high-quality datasets. While significant progress has been made in this area for major global currencies, there remains a notable gap in resources tailored to regional currencies, such as the Iranian Rial (IRR), Vietnamese Dong (VND) and Sri Lankan rupee (LKR).

Sri Lanka, a South Asian nation with a rich cultural and economic history, utilizes a unique set of currency notes and coins that reflect its heritage and economic development. The Sri Lankan rupee, subdivided into 100 cents, features distinct designs, texture, colors, and security features across its denominations. Recognizing these currencies accurately is particularly challenging due to variations in lighting conditions, wear and tear, and the presence of counterfeit notes. Despite these challenges, the lack of a comprehensive, publicly available dataset for Sri Lankan coins has limited the development of specialized coin recognition systems.

To address this gap, we introduce Sri Lankan CoinVision, a comprehensive image dataset specifically designed for currency recognition tasks. This dataset encompasses high-resolution images of all denominations of Sri Lankan coins, captured under diverse conditions to ensure robustness and generalizability. The dataset is annotated with detailed metadata, including denomination, currency type (coin or note), and image quality attributes, making it suitable for training and evaluating machine learning models. By introducing Sri Lankan CoinVision, we hope to bridge the gap in resources for regional currency recognition and contribute to the development of innovative solutions that address the unique challenges of the Sri Lankan financial ecosystem. This dataset represents a significant step forward in enabling researchers and developers to create more accurate, efficient, and inclusive currency recognition systems.

\section{Scientific Method Adapted for Coin Data Collection}\label{sec2}

To enhance the transparency of our data reporting, and to ensure the dataset meets the scientific standards, we followed a systematic and reproducible methodology for collecting and documenting coin data. The process involved the following steps:
\begin{enumerate}
	\item Selection of Coins
	\item  Imaging Methodology
	\item Metadata Documentation
	\item Data Validation and Quality Control
\end{enumerate}

A carefully curated collection of Sri Lankan coins was selected to encompass a diverse range of characteristics, ensuring a rich representation for various analytical purposes. The selection included newly minted coins, moderately used ones, and old or worn specimens, capturing the full spectrum of age-related variations. Coins in different conditions?clean, rusted, corroded, and damaged?were also considered to reflect real-world diversity. To enhance the breadth of the collection, multiple denominations were included, along with coins composed of various metals such as nickel, copper, and brass. This comprehensive assortment was designed to support computational and classification tasks in fields like computer vision, numismatics, and forensic analysis, providing valuable insights into the visual and material diversity of Sri Lankan currency.

To ensure high-quality, standardized images, all coin photographs were captured in a carefully controlled environment. A high-resolution digital camera with a fixed focal length was used, maintaining consistent lighting conditions for optimal clarity. Diffused lighting was employed to minimize shadows and reflections, while a neutral, non-reflective background helped eliminate distractions. Each coin was photographed from multiple standardized angles, including the obverse (front) view, reverse (back) view, and side view to highlight thickness and edge features. Additional tilted angles were used to capture surface texture and depth variations. To further enhance uniformity, every coin was positioned at a fixed distance from the camera, ensuring consistent image quality across the entire collection.

Each captured image was meticulously annotated with detailed metadata to ensure precise traceability and comprehensive analysis. Every coin was assigned a unique identification number, along with its denomination and year of minting. The physical condition was documented, ranging from new and slightly worn to heavily corroded specimens. Additionally, weight and dimensions were accurately measured using a precision scale and calipers to maintain consistency. Where applicable, environmental factors such as exposure to humidity or oxidation were also recorded. This structured metadata provides valuable insights, enabling researchers to correlate visual coin features with their physical attributes for applications in numismatics, computer vision, and forensic studies.

By following a structured, reproducible scientific methodology, this dataset provides valuable, high-quality coin images that can contribute to various research domains. The documentation ensures transparency and facilitates reproducibility in future studies.
	
\section{Related Works}\label{sec3}
Currency recognition has been a widely researched topic in the fields of computer vision, machine learning, and pattern recognition. Over the years, numerous studies have focused on developing automated systems for currency identification, driven by the need for efficient cash handling, counterfeit detection, and assistive technologies for visually impaired individuals. However, the development of robust systems heavily relies on the availability of comprehensive and high-quality datasets. While significant progress has been made for major global currencies, there remains a notable gap in resources tailored to regional currencies or currencies that have low exchange value against stronger global currencies like the US Dollar (USD), Euro (EUR), or British Pound (GBP). To enlighten the readers about different coin datasets available and the mechanisms used in currency recognition, this section reviews the relavent existing literature while emphasizing the gap in resources for regional currencies like the Sri Lankan rupee.

A recent and notable contribution to the field is the TurCoins dataset \cite{temiz2021turcoins}, 
which focuses on Turkish Republic coins minted from 1924 to 2019. This comprehensive dataset contains 11,080 coin images across 138 different classes, representing both the obverse and reverse sides of 5,540 distinct coins. The dataset encompasses 69 different main coin designs from Turkish Republic history, making it a valuable resource for numismatic research and coin classification tasks. To address the challenge of class imbalance, where some classes had over 300 samples while rare classes had as few as 20, the researchers developed a novel StyleGAN2-based synthetic coin generation method. This approach significantly improved classification accuracy for rare coin classes, increasing it from 37\% to 62\%, and outperformed traditional rotation-based augmentation by 2.3\%. The dataset is organized into three categories based on sample size: LOW (classes with $\leq$ 30 samples), MEDIUM (classes with 31-99 samples), and HIGH (classes with $\geq$ 100 samples). Experimental results demonstrated balanced performance across these categories when using equal samples per class. The researchers employed a ResNet50 model pre-trained on ImageNet and implemented transfer learning for coin classification, achieving an impressive 97.71\% accuracy with traditional data augmentation. 

The Ghana Currency Dataset (GC3558) is a comprehensive and openly accessible dataset designed to support classification modeling for currency recognition. It comprises 3,558 high-resolution color images spanning 13 distinct classes, including both coins and paper notes from Ghana. The dataset was meticulously curated by capturing images under diverse environmental conditions, such as white, dark, yellow, and illuminated backgrounds. This variability ensures robustness in training machine learning models, enabling them to perform well across different lighting and background scenarios. The dataset's creation process involved the use of high-resolution cameras, ensuring detailed and clear images suitable for advanced image processing tasks \cite{adu2022gc3558}. 

The Gupta Archer-Type Coins Dataset \cite{al2024novel} is a specialized collection focused on historical coinage, specifically coins from the Gupta Empire. This dataset contains 507 RGB color images across 18 distinct classes, each representing a unique type of Gupta archer coin. The images were sourced from verified private collections and auction houses, ensuring authenticity and high quality. When tested with deep learning models, the dataset demonstrated strong performance, achieving an accuracy of 89.0\% using the ResNet-50 architecture. 

The Indian Coin Image Dataset \cite{suryawanshi2024image} is a significant contribution to the field, specifically tailored for machine learning applications in currency recognition. This dataset contains 6,672 images representing 53 different classes of Indian coins, making it one of the most extensive collections of its kind. The images were captured under various conditions and backgrounds, with a strong emphasis on standardization and quality control. This ensures that the dataset is well-suited for training and evaluating machine learning models, particularly those aimed at recognizing and classifying Indian currency. Another important contribution in Indian currency recognition tasks is the \textit{Indian Currency Dataset}, which laid the foundation for basic currency recognition tasks. This dataset was later expanded by Srivastava's Indian Currency Notes dataset on Kaggle, which included more denominations and image variations. These datasets have been widely used for training and evaluating currency recognition systems, particularly for Indian currency.

Another recent addition to this field is the comprehensive dataset of damaged Indian banknotes, which addresses a critical gap in existing datasets. This dataset contains 5,125 images of damaged banknotes, categorized into eight classes representing both old and new denominations. By including damaged notes, this dataset enables the development of more robust and practical currency recognition systems that can handle real-world challenges, such as torn, folded, or stained currency  \cite{meshram2023comprehensive}. Going beyond, modern usable coins and currencies, some research have been carried out on ancient coin collections and classifications.

Ancient coin datasets play a crucial role in numismatic research and historical analysis. A significant contribution in this area is the Roman Imperial Denarii dataset, which contains over 650 specimens of Roman coins. These coins represent 100 different issues and span approximately 300 years (27 BC to 244 AD), covering 44 issuing authorities. The dataset includes coins in various conditions, such as Fine, Very Fine, and Extremely Fine, making it particularly useful for studying the effects of coin preservation on automatic recognition methods \cite{arandjelovic2020images}.

The prior works on currency recognition highlights the effectiveness of deep learning models, particularly for assisting visually impaired individuals. A notable study proposes a VGG-16-based system for recognizing seven Pakistani banknote denominations (10, 20, 50, 100, 500, 1000, and 5000 PKR) using a dataset of 4,713 images captured under various conditions, including folded, occluded, blurred, and diverse backgrounds. The model, fine-tuned with transfer learning and augmented with custom dense layers, achieved 99\% accuracy and high F1-scores (99\%) across all classes, outperforming prior models like YOLO-v3, AlexNet, and SVM. Despite its success, the study is limited by its focus on Pakistani banknotes and reliance on images from a single camera (iPhone XS), which restricts generalizability \cite{nasir2024pakistani}.
In the same direction, Thai Currency Recognition Using the Xception Model \cite{ahmad2024deep}, aimed to recognize five denominations of Thai currency (20, 50, 100, 500, and 1000 THB). Using a dataset of 3,600 images captured under diverse lighting, backgrounds, and conditions (e.g., folded or occluded notes), the Xception model achieved exceptional performance, with 99.5\% training accuracy, 99.8\% validation accuracy, and perfect precision, recall, and F1-scores (1.00) for all denominations. Although the study has been successful, its scope is restricted to Thai currency, and additional testing in real-world settings is necessary. Both studies demonstrate the potential of transfer learning and modern architectures like VGG-16 and Xception for currency recognition, while highlighting the importance of broader datasets, real-world applicability, and deployment considerations. Similarly, \cite{rafi2024nstu} introduced the NSTU-BDTAKA dataset, which includes a detection subset (3,111 annotated images) and a recognition subset (28,875 images across nine denominations in Bangladeshi currency: 2, 5, 10, 20, 50, 100, 200, 500, and 1000 BDT). The YOLOv5 model was used for real-time detection, while recognition models were trained on labeled classes with extensive data augmentation (rotation, flipping, shearing, blurring). The study highlights reliable real-time detection and robust recognition capabilities but faces challenges such as inconsistent image quality, occlusion, and varying illumination. These three studies \cite{nasir2024pakistani, ahmad2024deep, rafi2024nstu} highlight pre-trained architectures (VGG-16, Xception, YOLOv5) and emphasize the importance of diverse datasets to handle real-world conditions. However, they share limitations in generalizability due to their focus on single-country currencies and reliance on limited hardware for image capture.

Despite the significant progress in the development of currency and coin dataset development, there remains a notable gap in the availability of a comprehensive dataset for Sri Lankan coins. While datasets like the Ghana Currency Dataset, Indian Currency Dataset, and TurCoins have advanced research in currency recognition and classification, no equivalent dataset exists for Sri Lankan coins. This gap limits the development of machine learning models and automated systems tailored to Sri Lankan currency, which is essential for applications such as numismatic research, historical analysis, and automated coin sorting.

The \texttt{Table} \ref{tab:table1}, summarizes the recent work regards to coin recognition tasks.

\FloatBarrier
\begin{table*}[!h]
	\scriptsize
	\caption{Summary - literature on coin classification}
	\centering
	\renewcommand{\arraystretch}{2} % Adjust row height
	\begin{tabular}{|p{1.25cm}|c|p{1.31cm}|c|p{4cm}|p{5cm}|}
		\hline
		\textbf{Paper} & \textbf{Sample} & \textbf{Country} & \textbf{Year} & \textbf{Data set} & \textbf{Specialty} \\
		\hline
		\cite{arandjelovic2020images} & \raisebox{-1.5\height}{\includegraphics[width=2cm]{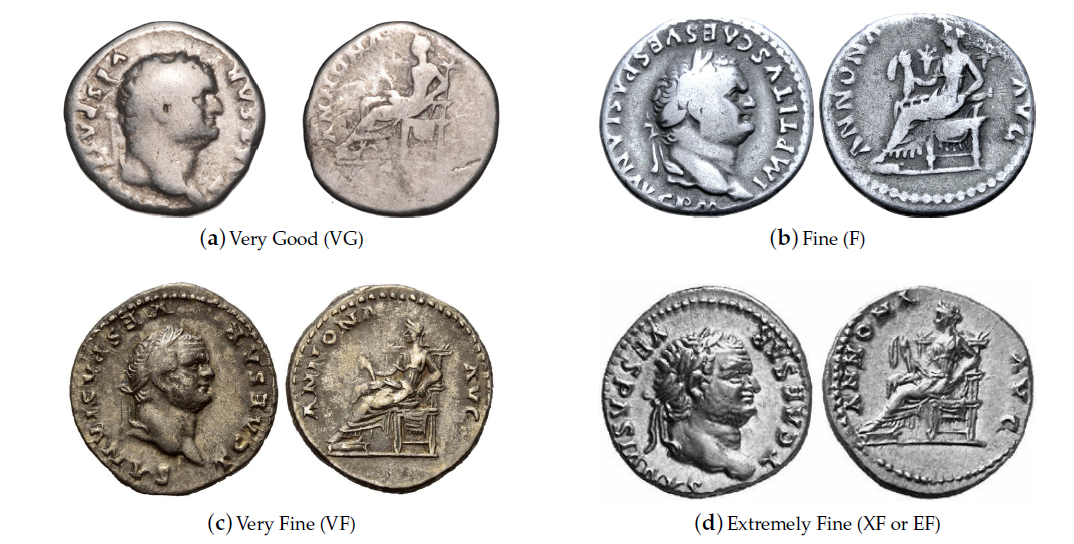}} & Ancient Rome (Italy) & 2020 & 650 specimens of Roman coins, include coins in various conditions, such as Fine, Very Fine, and Extremely Fine & 100 different issues and span approximately 300 years \\
		\hline
		\cite{temiz2021turcoins} &
		\raisebox{-1.5\height}{\includegraphics[width=2cm]{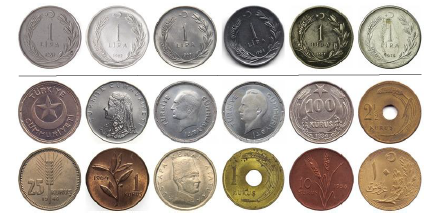}}
		& Turkey & 2021 & 11,080 images across 138 classes, representing
		both the obverse and reverse sides of 5,540 distinct coins.
		69 different main coin designs from
		Turkish Republic history & StyleGAN2-based synthetic coin generation
		method\\
		\hline
		\cite{adu2022gc3558} & \raisebox{-1.5\height}{\includegraphics[width=2cm]{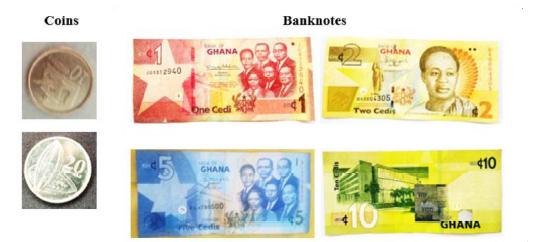}} & Ghana & 2022 & 3,558 high-
		resolution color images of 13 distinct classes, both coins and paper notes. &  use of high-resolution
		cameras, capturing images under diverse
		environmental conditions, such as white, dark, yellow, and
		illuminated backgrounds\\
		\hline
		\cite{meshram2023comprehensive} & \raisebox{-1.5\height}{\includegraphics[width=2cm]{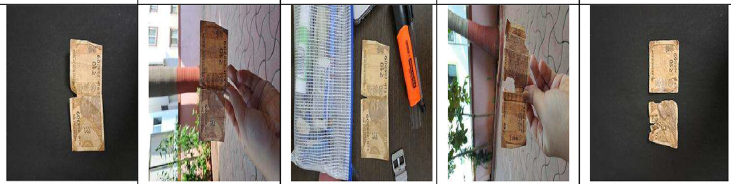}} & India & 2023 & 5,125 images of damaged banknotes, categorized into 08 classes & comprehensive data-set of damaged Indian banknotes \\
		\hline
		\cite{al2024novel} & \raisebox{-2.5\height}{\includegraphics[width=2cm]{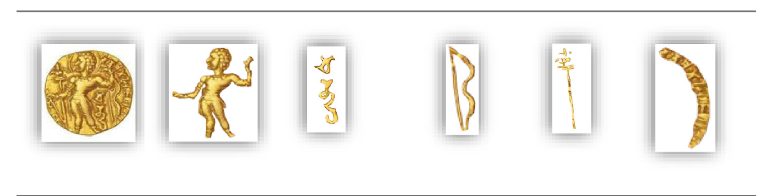}} & India (Gupta)& 2024 & 507 RGB
		color images across 18 distinct classes & Coins from verified private collections and auction houses, ensuring authenticity and high quality. With ResNet-50 architecture, the dataset demonstrated strong performance, achieving an accuracy of 89.0\%.  \\
		\hline
		\cite{suryawanshi2024image}  & \raisebox{-1.5\height}{\includegraphics[width=2cm]{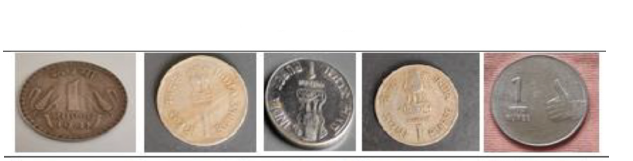}} & India & 2024 & 6,672 images representing 53 different classes, captured under various conditions and backgrounds & dateset was later expanded by Srivastava's Indian Currency Notes dateset on Kaggle, which included more denominations and image variations \\
		
		\hline
		\cite{ahmad2024deep} & \raisebox{-1.5\height}{\includegraphics[width=1.5cm]{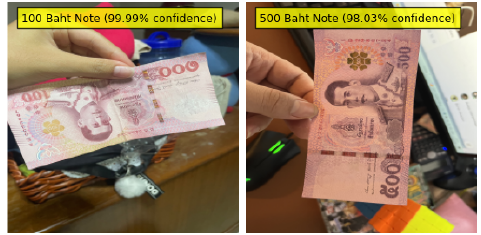}} & Thailand & 2024 & dataset of 3,600 images captured under diverse lighting, backgrounds
		& Xception model achieved exceptional performance, with 99.5\% training accuracy, 99.8\% validation accuracy
		\\
		
		\hline
		\cite{nasir2024pakistani} & \raisebox{-1.5\height}{\includegraphics[width=1.5cm]{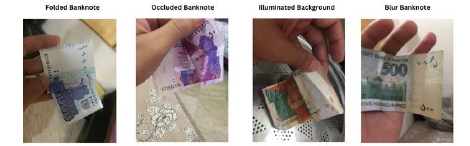}} & Pakistan & 2024 & 4,713 images captured under various conditions, including folded, occluded, blurred
		& transfer learning and augmented with custom dense layers, achieved 99\% accuracy and high F1-scores (99\%) across all classes
		\\
		
		\hline
		\cite{rafi2024nstu}  & \raisebox{-1.5\height}{\includegraphics[width=1.5cm]{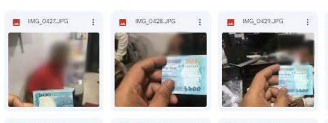}} & Bangladesh & 2024 & detection subset (3,111 annotated images) and a recognition subset (28,875 images across nine denominations: 2, 5, 10, 20, 50, 100, 200, 500, and 1000 BDT)
		
		& YOLOv5 model was used. reliable real-time detection and robust recognition capabilities. challenges such as inconsistent image quality, occlusion, and varying illumination.
		\\
		\hline
	\end{tabular}
	\label{tab:table1}
\end{table*}
\FloatBarrier

\section{Methodology}\label{sec4}
The methodology adopted in this study follows a well-structured and systematic approach to ensure the development of a comprehensive and representative dataset specifically for the classification of Sri Lankan coins. This methodology is designed to enhance the accuracy, efficiency, and generalizability of the model by incorporating various essential stages, including data collection, image pre-processing, dataset structuring, and data augmentation. Each of these steps plays a critical role in refining the dataset and optimizing the performance of the classification models.

The initial phase of the study focuses on systematic data collection, ensuring that a diverse and well-balanced dataset is gathered. This dataset includes images of different Sri Lankan coin denominations under various lighting conditions, orientations, and backgrounds to create a robust and representative sample. Following data collection, image pre-processing techniques are applied to standardize the dataset, enhancing the quality of the images while reducing noise and inconsistencies. This process involves resizing, grayscale conversion, contrast adjustment, and background removal to prepare the images for further analysis.

To improve the model's ability to generalize across varying conditions, data augmentation techniques are employed. These techniques include transformations such as rotation, scaling, flipping, and brightness adjustments, ensuring that the model can recognize coins under different scenarios and real-world conditions. By diversifying the training data, augmentation enhances the model?s robustness and reduces the risk of overfitting.

The core of the study involves evaluating multiple feature extraction techniques and classification models to identify the most effective approach for recognizing and classifying Sri Lankan coins. The study is structured into three distinct levels of analysis. At the first level, traditional feature extraction techniques are explored using OpenCV?s Oriented FAST and Rotated BRIEF (ORB) algorithm. This approach focuses on detecting key points and descriptors to differentiate between various coin denominations based on their unique patterns and textures.

At the second level, deep learning-based methods are integrated to enhance classification performance. MobileNetV2, a lightweight convolutional neural network (CNN) architecture known for its efficiency in mobile and embedded vision applications, is employed for feature extraction and classification. MobileNetV2 is particularly advantageous due to its ability to balance computational efficiency with high accuracy, making it a suitable choice for coin classification tasks.

The third and final level of analysis involves the development of a custom Convolutional Neural Network (CNN) architecture tailored specifically for Sri Lankan coin classification. This custom CNN is designed with optimized layers and parameters to extract intricate features from coin images, thereby improving classification accuracy. By leveraging deep learning techniques, the model can learn complex patterns and features that may not be easily captured by traditional feature extraction methods.

Through this multi-stage evaluation, the study aims to compare the effectiveness of different methodologies, ultimately determining the most efficient and accurate approach for Sri Lankan coin classification. By integrating traditional computer vision techniques with modern deep learning architectures, this research seeks to establish a reliable framework for automated coin recognition, which can be beneficial for various applications, including numismatics, vending machines, and digital payment systems.
\subsection{Dataset Creation}

\begin{figure}[h!]
	\caption{Coin Dataset Creation Process}
	\centering
	\includegraphics[width=0.8\linewidth]{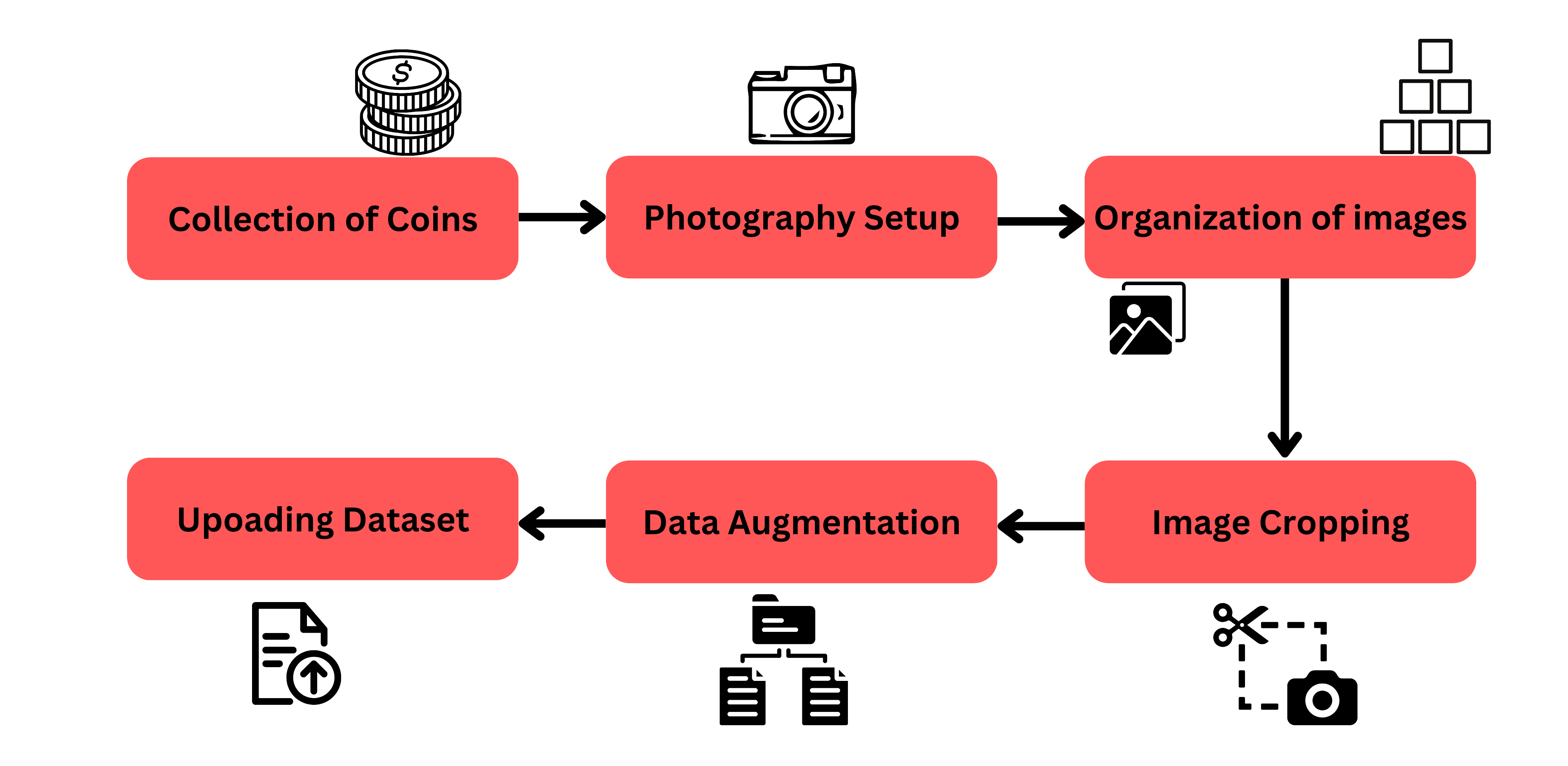}
	\label{fig:coin-dataset}
\end{figure}

The methodology for this study involves a series of structured steps to ensure the creation of a high-quality and representative dataset. As illustrated in Fig. \ref{fig:coin-dataset}, the process begins with the collection of coins in various conditions, followed by systematic image capture, organization, preprocessing, and augmentation.

%methodology Diagram
\FloatBarrier
\FloatBarrier

\subsubsection{Collection of Coins} 
\begin{figure}[!h]
	\caption{Coin Condition Variations}
	\centering
	% First Image
	\begin{minipage}{0.20\textwidth} % Adjust the width of each image
		\includegraphics[width=\textwidth]{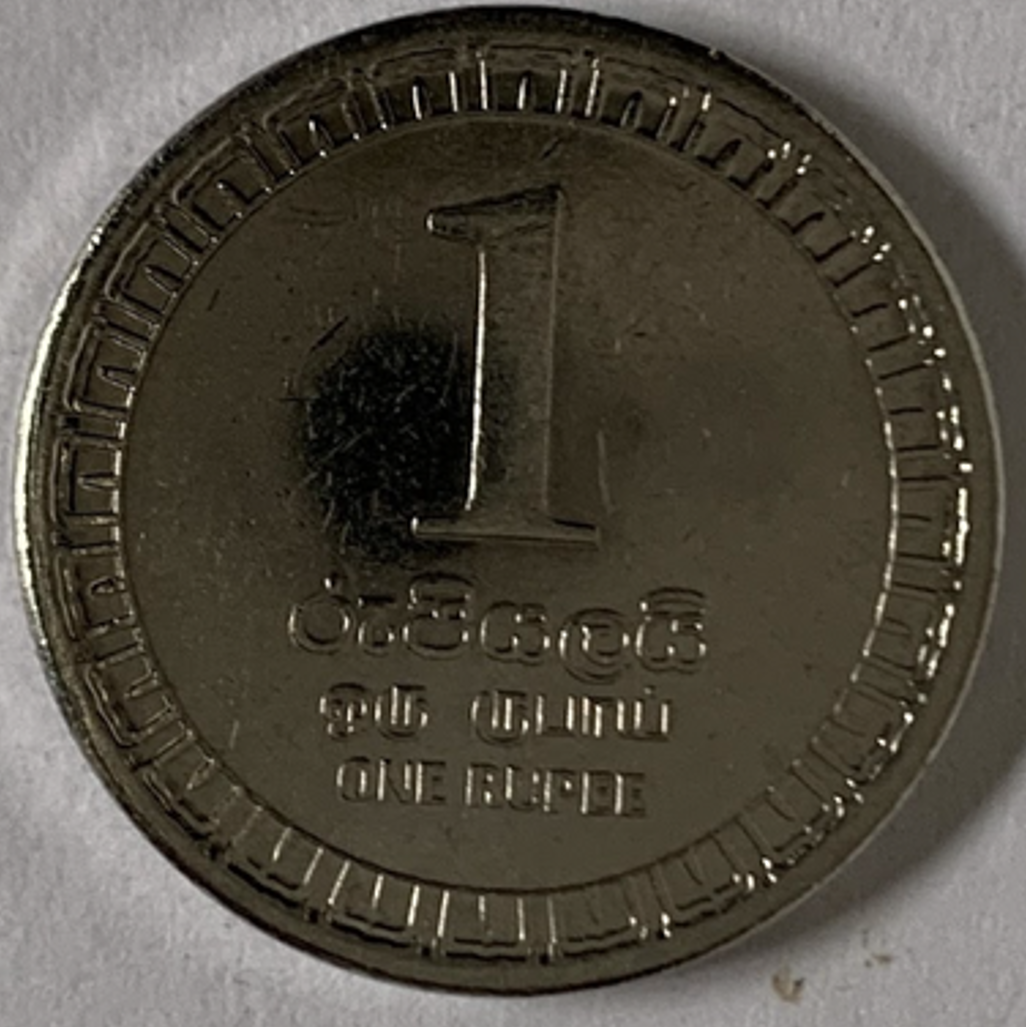}
		% \caption*{Image 1} % Optional caption for each image
	\end{minipage}
	\hspace{0.005\textwidth} % Horizontal space between images
	% Second Image
	\begin{minipage}{0.20\textwidth}
		\includegraphics[width=\textwidth]{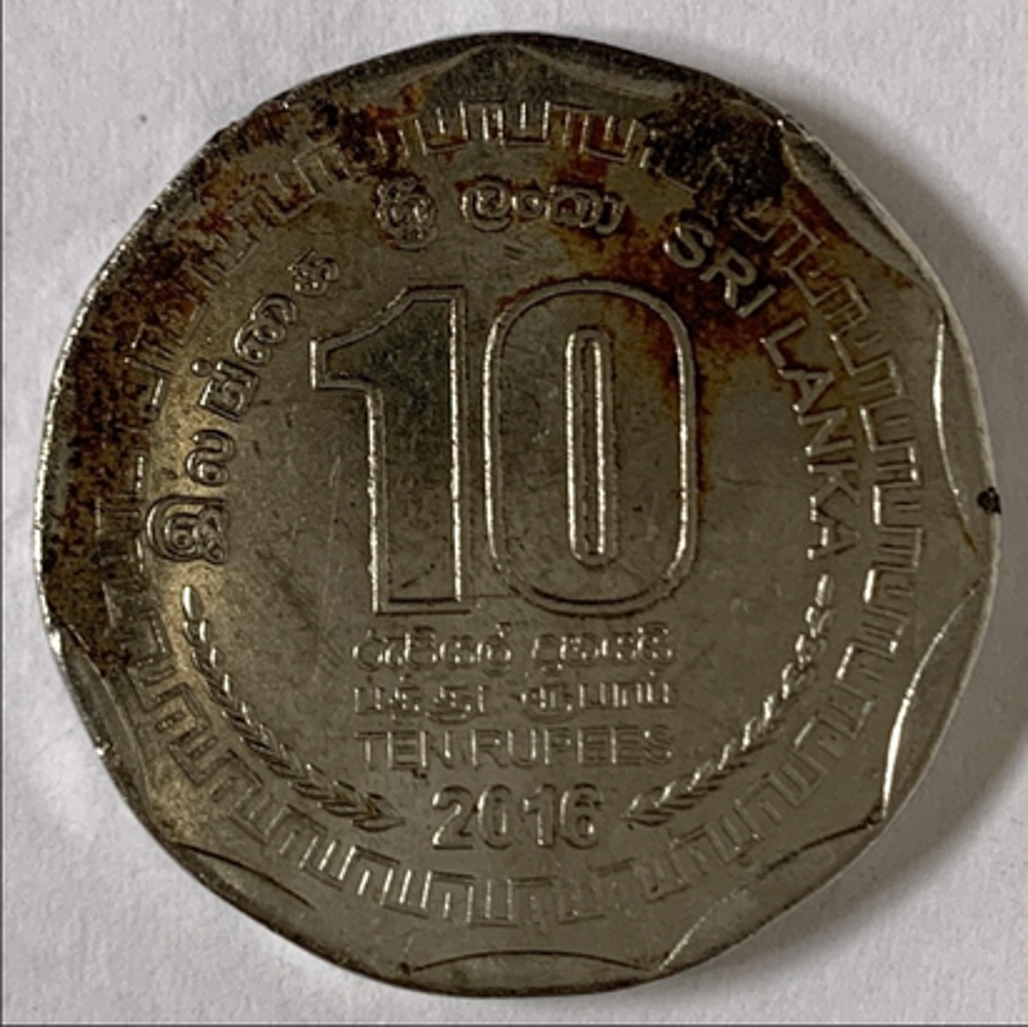}
		% \caption*{Image 2}
	\end{minipage}
	\hspace{0.005\textwidth}
	% Third Image
	\begin{minipage}{0.20\textwidth}
		\includegraphics[width=\textwidth]{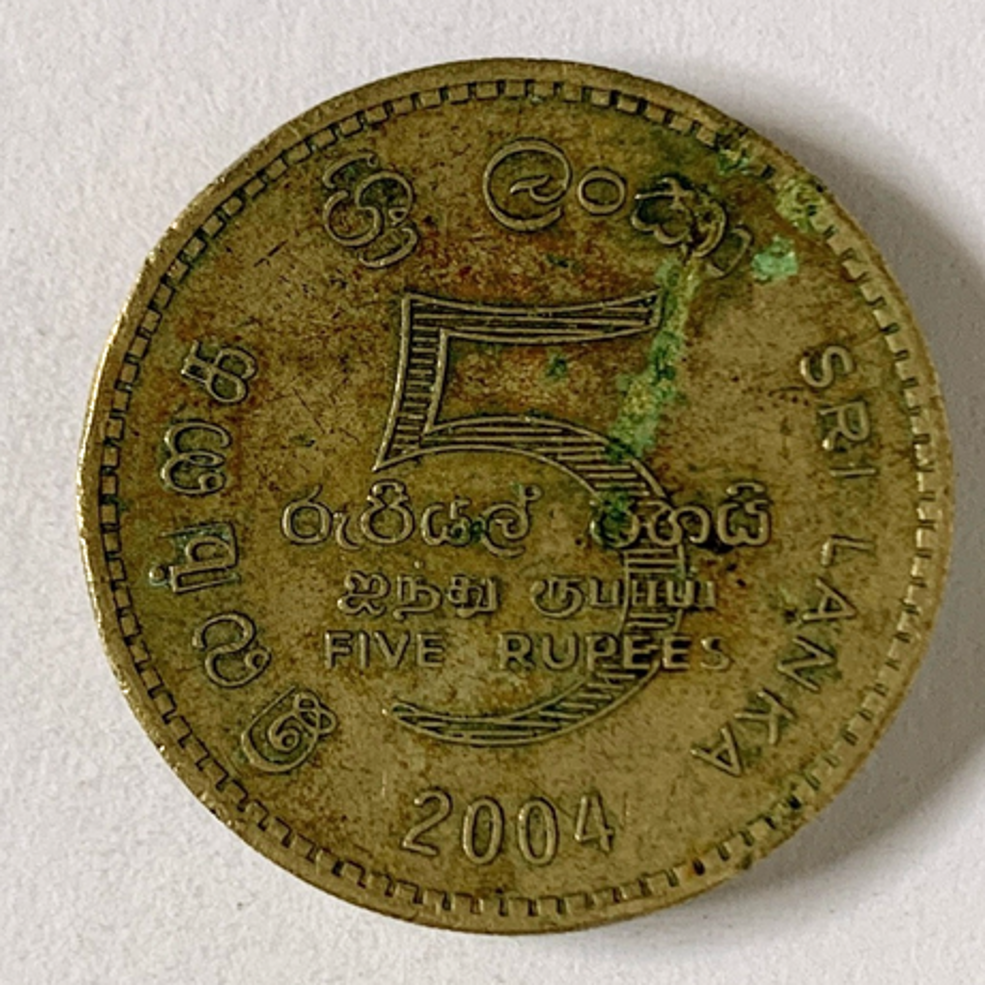}
		% \caption*{Image 3}
	\end{minipage}
	\hspace{0.005\textwidth}
	% Fourth Image
	\begin{minipage}{0.20\textwidth}
		\includegraphics[width=\textwidth]{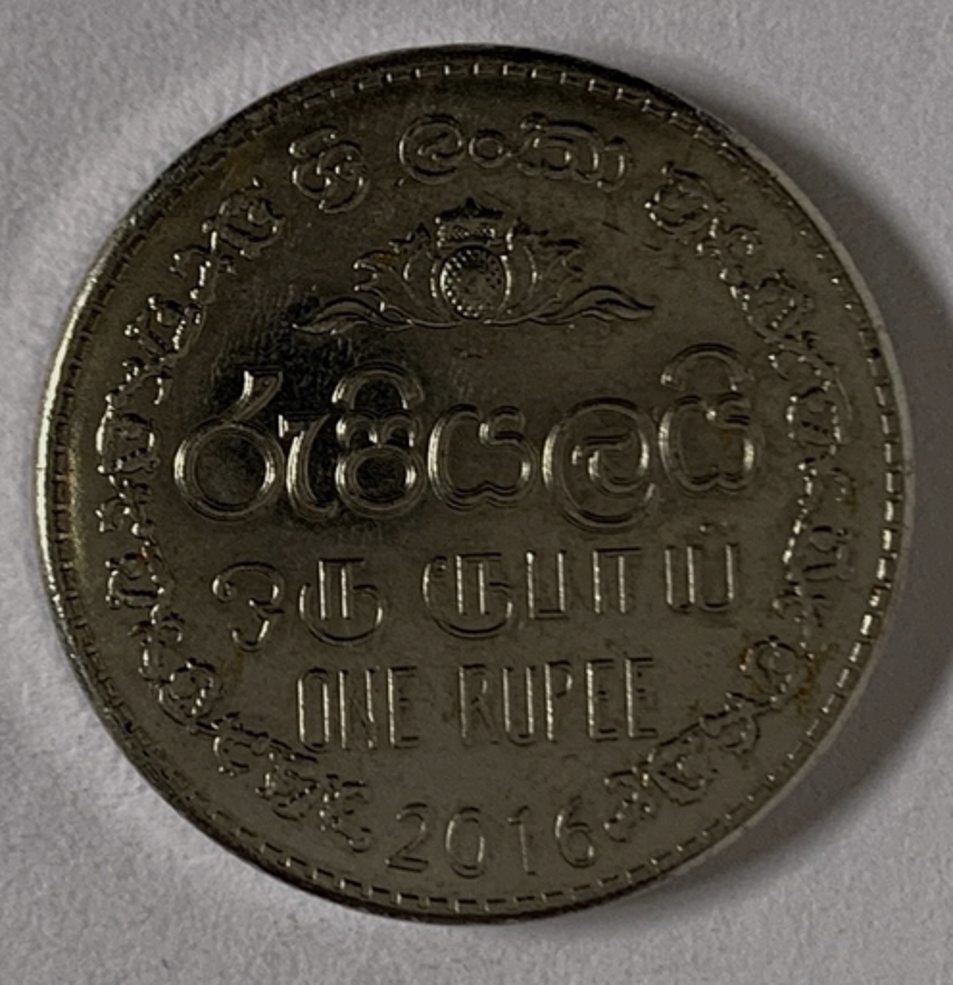}
		% \caption*{Image 4}
	\end{minipage}

	\label{fig:horizontal-images}
\end{figure}

The process of creating the dataset began with the collection of coins from all currently circulating denominations. The dataset was structured around Sri Lankan coins issued by the Central Bank of Sri Lanka (CBSL) \cite{cbsl_coins}, incorporating both new and old versions of the currency. The dataset comprises four main denominations of Sri Lankan coins, each with two versions: old and new. These include Rs. 10 (old and new versions), Rs. 5 (old and new versions), Rs. 2 (old and new versions), and Rs. 1 (old and new versions). To ensure the dataset was diverse and representative of real-world scenarios, coins in various conditions were gathered. According to the Fig. \ref{fig:horizontal-images}, these included coins that were rusted, dented, or otherwise worn, as well as those in relatively new or pristine conditions.
\FloatBarrier

\FloatBarrier

\subsubsection{Photography Setup} 
High-quality images of the collected coins were captured using a 12 MP camera with advanced specifications, including an f/1.8 aperture, a 26mm wide lens, and optical image stabilization (OIS). To maintain consistency across all images, the photos were taken from a fixed distance under controlled lighting conditions. A 1:1 aspect ratio was used to ensure uniformity in the image dimensions. However, due to the limitations of the camera?s focus capabilities at close range, the coins did not occupy the entire frame in the photos. This was a deliberate choice to avoid losing image clarity, as getting too close to the coins would result in blurred or unfocused images. The consistent setup ensured that lighting, angle, and focus remained uniform across the dataset.

\subsubsection{Organization of Images}  
Once the images were captured, they were systematically organized into distinct classes based on the coin denominations and their physical conditions. This step involved categorizing the coins into groups such as denomination type (e.g., rs1, rs2) and condition (e.g., new, old, rusted, dented). Organizing the images into classes was crucial for creating a structured dataset that could be easily used for training and evaluation purposes. 

\subsubsection{Image Cropping} 
Since the coins did not fill the entire frame in the captured images, manual cropping was performed to isolate the coins and make them the central focus of each image. This step was necessary because the physical size of the coins varied significantly. For example, some coins, like the rs1version2 (new), were much smaller, while others, like the rs2 (old and new), were larger. Cropping was done manually to ensure precision, as automated cropping tools might not accurately account for the size differences. 
\subsubsection{Data Augmentation} 
\begin{figure}[!h]
	\caption{Data augmentation techniques}
	\centering
	% First Image
	\begin{subfigure}[b]{0.2\textwidth} % Adjust the width of each image
		\includegraphics[width=\textwidth]{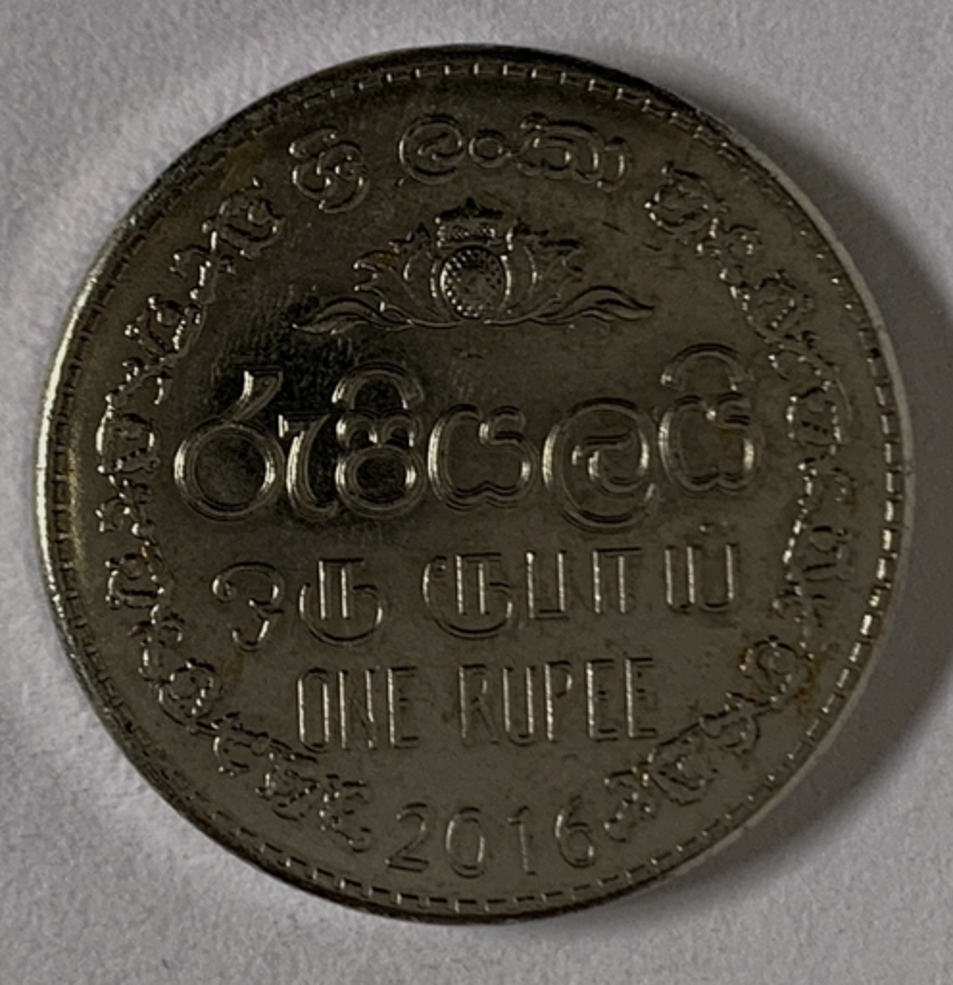}
		\caption{Zoom In} % Subcaption for this image
	\end{subfigure}
	\hspace{0.01\textwidth} % Horizontal space between images
	% Second Image
	\begin{subfigure}[b]{0.2\textwidth}
		\includegraphics[width=\textwidth]{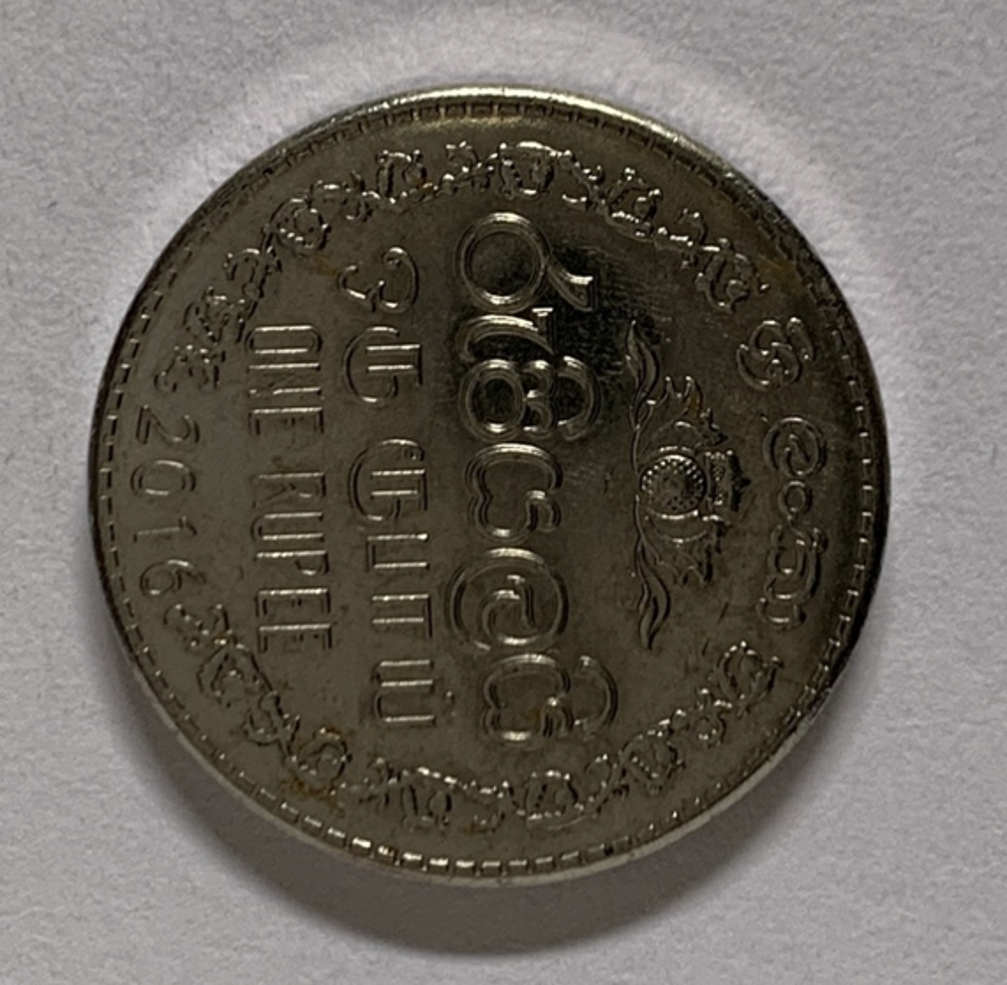}
		\caption{Zoom Out}
	\end{subfigure}
	\hspace{0.01\textwidth}
	% Third Image
	\begin{subfigure}[b]{0.2\textwidth}
		\includegraphics[width=\textwidth]{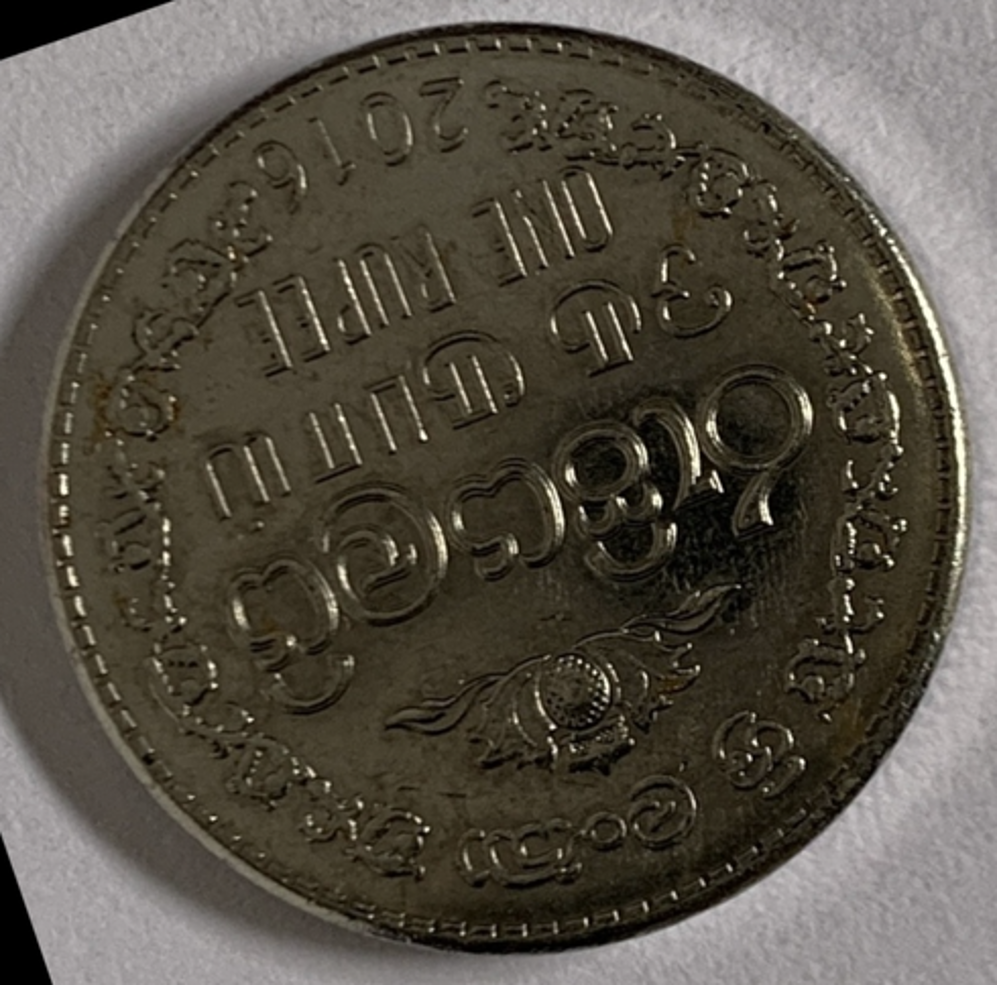}
		\caption{Rotation}
	\end{subfigure}
	\hspace{0.01\textwidth}
	% Fourth Image
	\begin{subfigure}[b]{0.2\textwidth}
		\includegraphics[width=\textwidth]{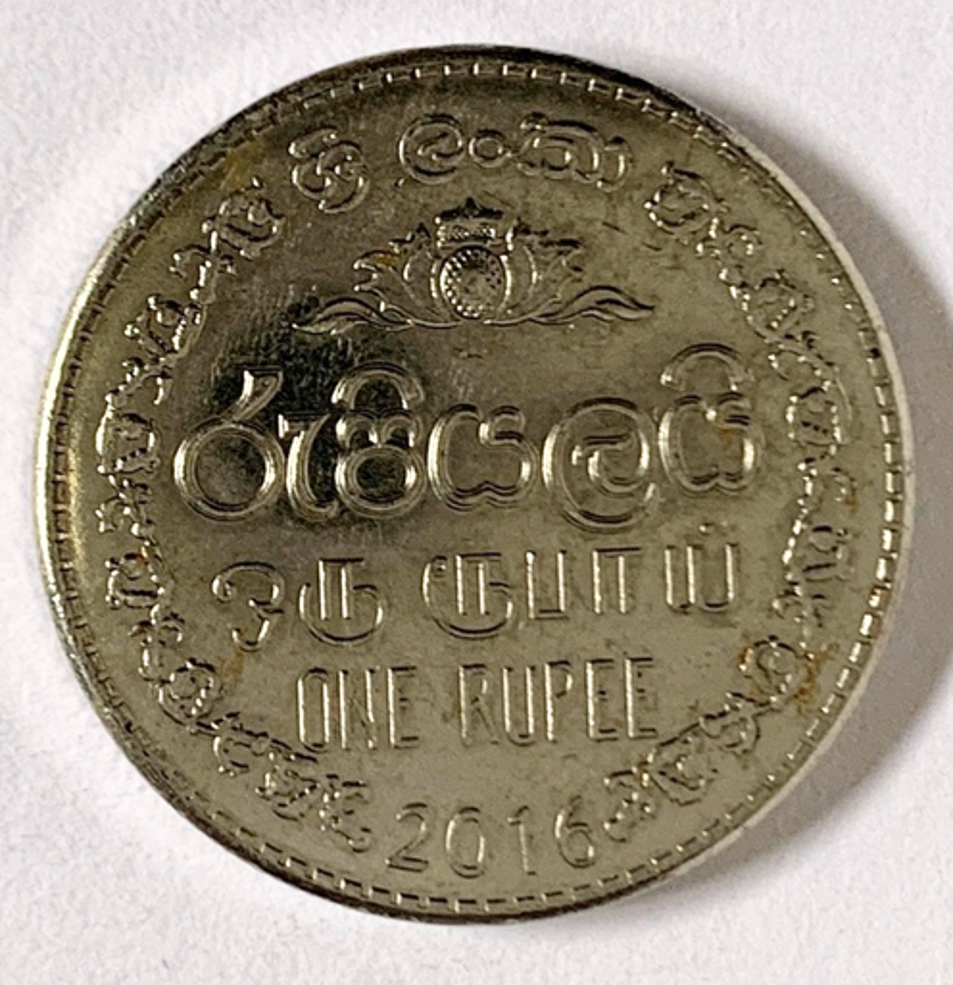}
		\caption{Brightness Adj.}
	\end{subfigure}

	\label{fig:augmentation}
\end{figure}

To further enhance the dataset?s validity and prepare it for real-world applications, several data augmentation techniques were applied. Fig. \ref{fig:augmentation} illustrates the transformations. Random brightness adjustments were introduced to simulate varying lighting conditions, such as bright daylight or dim indoor lighting. Random zooming was used to create slight out-of-focus effects and minor cropping, mimicking scenarios where coins might not be perfectly centered or focused in an image. Additionally, the images were rotated at intervals of 20 degrees to account for different orientations. This was particularly important for coins with text or letters, as their recognition could vary depending on the angle at which they were viewed. 

\FloatBarrier
\FloatBarrier

\subsubsection{Dataset Organization}
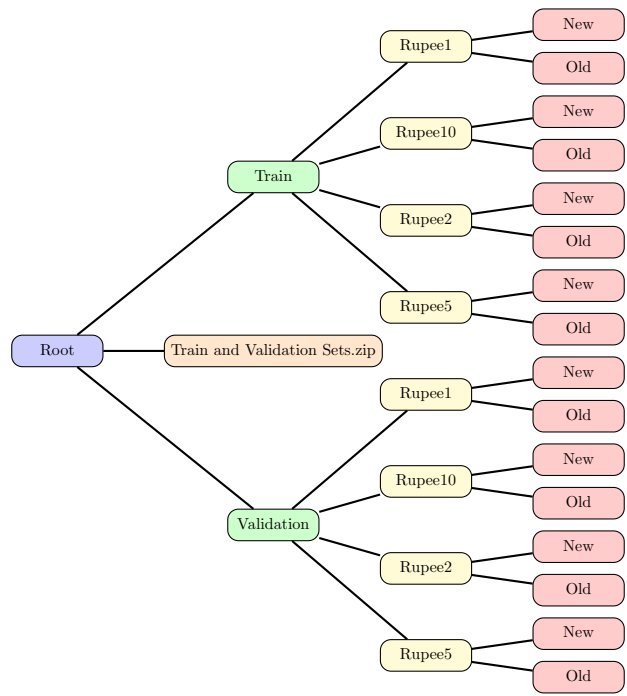
\begin{figure}[!h]
	\caption{Directory Structure of the Dataset}
	\label{fig:dataset-structure}
	\centering
	\begin{forest}
		for tree={
			scale=0.6,
			grow'=0, % Horizontal layout
			draw, % Draw rectangles for nodes
			rounded corners, % Rounded edges for nodes
			minimum width=2cm, % Set equal width for all rectangles
			minimum height=0.5cm, % Set equal height for all rectangles
			align=center, % Center-align text inside rectangles
			l sep=8mm, % Vertical spacing between levels
			s sep=1.5mm, % Horizontal spacing between sibling nodes
			edge={draw=black, thick}, % Black lines connecting nodes
			font=\footnotesize, % Font size of text
			text centered % Center text in all nodes
		}
		% Tree structure begins
		[Root, fill=blue!20
		[Train, fill=green!20
		[Rupee1, fill=yellow!20
		[New, fill=red!20]
		[Old, fill=red!20]
		]
		[Rupee10, fill=yellow!20
		[New, fill=red!20]
		[Old, fill=red!20]
		]
		[Rupee2, fill=yellow!20
		[New, fill=red!20]
		[Old, fill=red!20]
		]
		[Rupee5, fill=yellow!20
		[New, fill=red!20]
		[Old, fill=red!20]
		]
		]
		[Train and Validation Sets.zip, fill=orange!20]
		[Validation, fill=green!20
		[Rupee1, fill=yellow!20
		[New, fill=red!20]
		[Old, fill=red!20]
		]
		[Rupee10, fill=yellow!20
		[New, fill=red!20]
		[Old, fill=red!20]
		]
		[Rupee2, fill=yellow!20
		[New, fill=red!20]
		[Old, fill=red!20]
		]
		[Rupee5, fill=yellow!20
		[New, fill=red!20]
		[Old, fill=red!20]
		]
		]
		]
	\end{forest}
	
\end{figure}

According to the Fig. \ref{fig:dataset-structure}, this dataset is organized into two main directories: `Train' and `Validation'.  
Each directory contains subfolders corresponding to different denominations of Rupee coins, categorized into `New' and `Old' versions.  Additionally, a compressed file named `Train and Validation Sets.zip' is included, which contains the dataset in a compact form.

\subsection{Model Training}
In this study, we adopted a three-level approach to evaluate the effectiveness of different feature extraction and modeling techniques for Sri Lankan coin classification, ensuring a thorough examination of our dataset's performance across various methods.

\subsubsection{Level 1 - with OpenCV ORB} We used OpenCV ORB (Oriented FAST and Rotated BRIEF) for feature extraction, generating keypoints and descriptors that were converted into fixed-size feature vectors. These features were then used to train three classifiers: K-Nearest Neighbors (KNN), Random Forest (RF), and Support Vector Machine (SVM), with their performance evaluated using metrics such as accuracy, precision, recall, and F1-score. 

\subsubsection{Level 2 - with MobileNetV2}
We employed MobileNetV2, a pre-trained deep learning model, as a feature extractor by removing its classification head and extracting features from the last convolutional layer. These features were used to train the same classifiers (KNN, RF, SVM). The performance evaluated using metrics such as accuracy, precision, recall, and F1-score.
\FloatBarrier

\FloatBarrier

\subsubsection{Level 3 - with a cutom CNN} We designed a custom convolutional neural network (CNN) accorsing to the Fig. \ref{CNN}, from scratch, incorporating convolutional layers, fully connected layers, and dropout for regularization. The custom CNN accepts input images of shape (128, 128, 3) and consists of three convolutional blocks, each followed by max pooling layers. The convolutional layers use 32, 64, and 128 filters, respectively, to extract hierarchical features. The output from the final convolutional layer is flattened into a 1D vector and passed through two dense layers, with dropout applied to prevent over-fitting. The final dense layer outputs 8 neurons, corresponding to the 8 coin classes. The custom CNN was trained on the Sri Lankan coin dataset, and its performance was evaluated and compared with the previous levels. 

Finally, we conducted a comprehensive analysis of the results, to identify the most effective approach for Sri Lankan coin classification. This structured methodology allowed us to systematically evaluate the utility of traditional and deep learning-based techniques for the task, ensuring that our dataset's performance was rigorously tested across different feature extraction methods, classifiers, and a custom model.
\FloatBarrier
\begin{figure}[!h]
	\caption{CNN Model Architecture}
	\centering
	%\hspace*{-2.8cm}
	\includegraphics[scale=0.34]{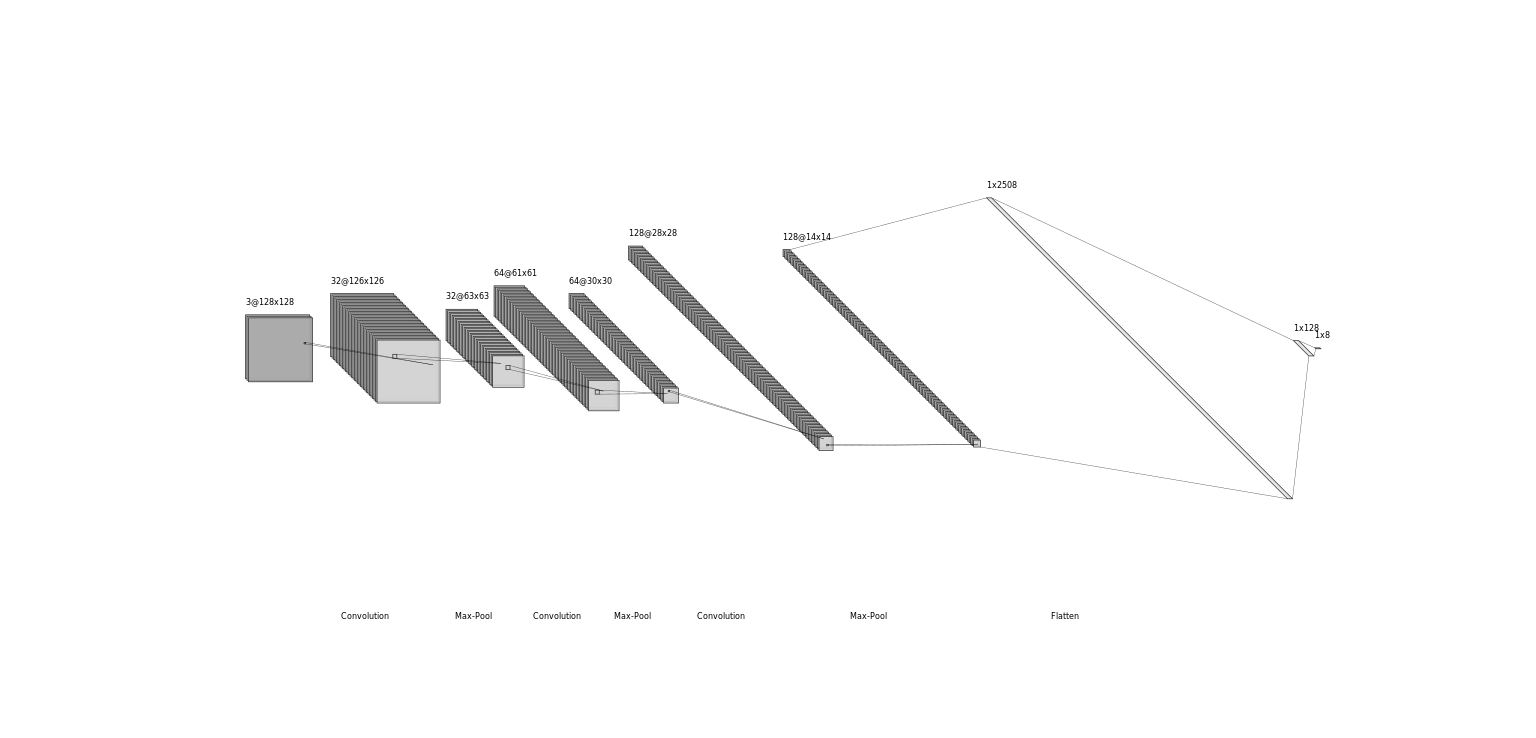}
	\label{CNN}
\end{figure}
\FloatBarrier
\begin{figure}[!h]
	%\caption{CNN Model Architecture}
	\centering
	\hspace*{1cm}
	\includegraphics[scale=0.25]{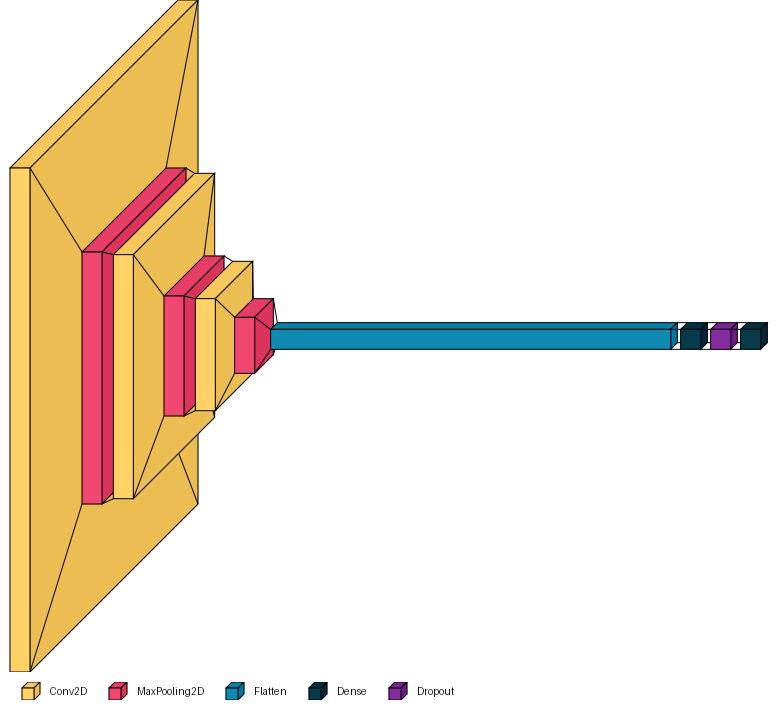}
	
	\label{CNN}
\end{figure}
\FloatBarrier
\section{Results and Discussion}

The Sri Lankan Coins Image Dataset is publicly accessible on Kaggle \cite{kaggle_dataset_sri_lankan_coins}. Table \ref{tab:denomination_images} displays the respective image counts for each coin. 

\begin{figure}[!h]
	\caption{Level 1 Results}
	\centering
	% First Image
	\begin{subfigure}[b]{0.23\textwidth} % Adjust the width of each image
		\includegraphics[width=\textwidth]{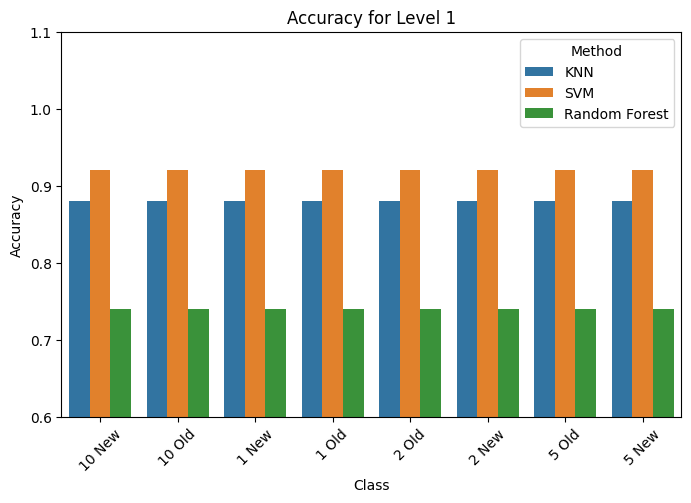}
		\caption{Accuracy} % Subcaption for this image
	\end{subfigure}
	\hspace{0.01\textwidth} % Horizontal space between images
	% Second Image
	\begin{subfigure}[b]{0.23\textwidth}
		\includegraphics[width=\textwidth]{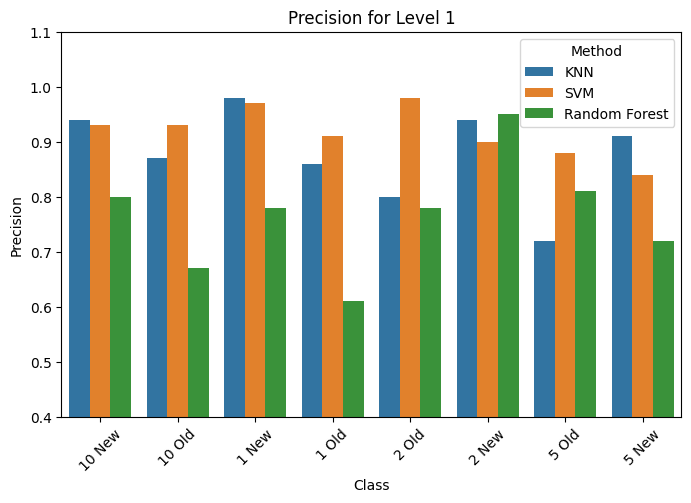}
		\caption{Precision}
	\end{subfigure}
	\hspace{0.01\textwidth}
	% Third Image
	\begin{subfigure}[b]{0.23\textwidth}
		\includegraphics[width=\textwidth]{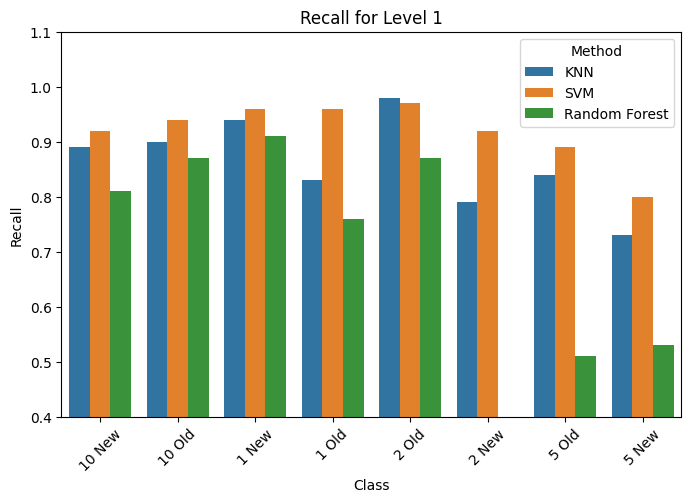}
		\caption{Recall}
	\end{subfigure}
	\hspace{0.01\textwidth}
	% Fourth Image
	\begin{subfigure}[b]{0.23\textwidth}
		\includegraphics[width=\textwidth]{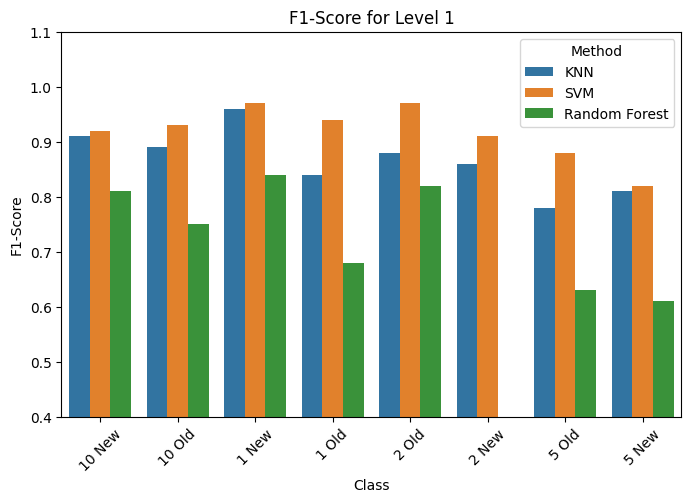}
		\caption{F1-Score}
	\end{subfigure}

	\label{fig:L1_results}
\end{figure}

In the landscape of machine learning datasets, recent advancements have yielded significant contributions tailored for machine learning applications. We aim to demonstrate the utility of the Sri Lankan coin image dataset through experiments conducted with established classifiers, namely K-Nearest Neighbors (KNN), Support Vector Machines (SVM), and Random Forest. The primary objective is to evaluate the dataset?s impact on enhancing the accuracy of machine learning models, particularly in the identification of Sri Lankan coins. Initially, the classifiers were applied without modifications, serving as a benchmark against our dataset.
\FloatBarrier
\begin{table}[!h]
	\caption{Denomination-wise number of images}
	\centering
	\renewcommand{\arraystretch}{1.5}
	\begin{tabular}{|c|c|}
		\hline
		\textbf{Denomination} & \textbf{Number of Images} \\
		\hline
		One Rupee (Old) & 3857 \\ \hline
		One Rupee (New) & 1672 \\ \hline
		Two Rupee (Old) & 2565 \\ \hline
		Two Rupee (New) & 1919 \\ \hline
		Five Rupee (Old) & 3059 \\ \hline
		Five Rupee (New) & 4693 \\ \hline
		Ten Rupee (Old) & 3477 \\ \hline
		Ten Rupee (New) & 2565 \\
		\hline
	\end{tabular}
	\label{tab:denomination_images}
\end{table}
\FloatBarrier
The Level 1 implementation results in Fig. \ref{fig:L1_results} demonstrate distinct performance patterns across the three machine learning models. The accuracy graph shows SVM consistently maintaining superior performance around 0.92 across all coin classes, while KNN achieves moderate accuracy of approximately 0.88, and Random Forest trails with roughly 0.74 accuracy. In terms of precision, both KNN and SVM exhibit strong performance, particularly for Rs1-New coins reaching 0.98, though Random Forest shows inconsistent precision with notably low performance for Rs1-Old at 0.61. The recall metrics reveal SVM's robust performance between 0.92-0.96, with KNN showing particularly strong recall for Rs2-Old at 0.98, while Random Forest struggles significantly with Rs5 denominations, achieving only 0.51-0.53 recall. The F1-score graph synthesizes these metrics, highlighting SVM's consistent excellence (0.91-0.97), KNN's solid performance (0.84-0.96), and Random Forest's challenges, particularly with Rs2-New where it achieves only 0.28. These visualizations clearly establish SVM as the most reliable classifier for Sri Lankan coin recognition in Level 1 implementation, demonstrating superior performance across all evaluation metrics.
\FloatBarrier

\begin{center}
	\begin{figure}[!h]
		\centering
		\includegraphics[width=0.45\textwidth]{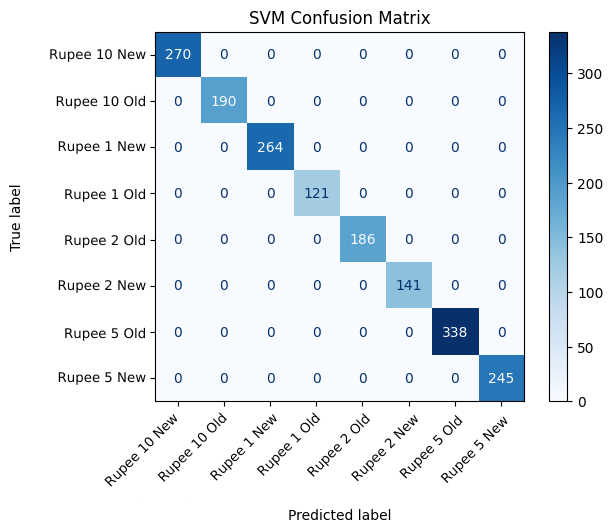}
		\caption{Confusion matrix of SVM (Level 2)}
		\label{Conf_L2_SVM} 
	\end{figure}
\end{center}

\begin{figure}[!h]
	\centering
	\includegraphics[width=0.45\textwidth]{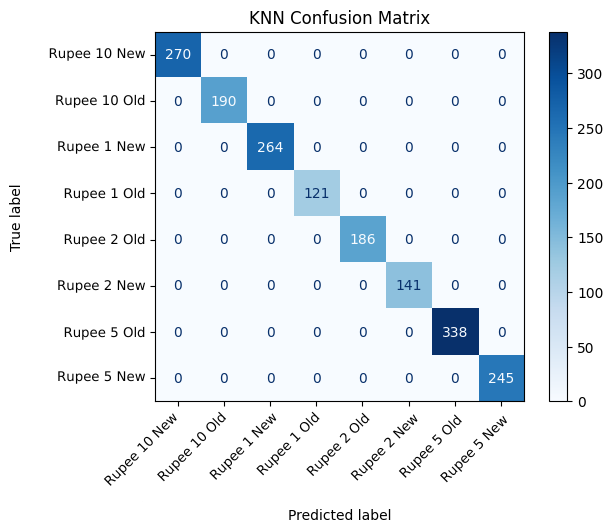}
	\caption{Confusion matrix of KNN (Level 2) } 
	\label{Conf_L2_KNN}
\end{figure}

\begin{figure}[!h]
	\centering
	\includegraphics[width=0.5\textwidth]{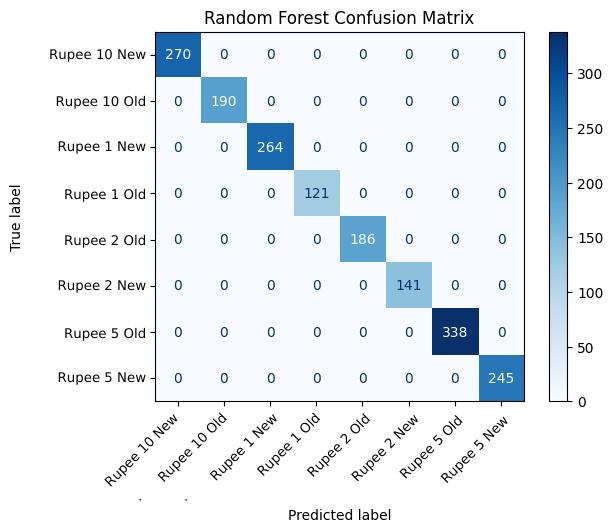}
	\caption{Confusion matrix of RF (Level 2)}
	\label{Conf_L2_RF}
\end{figure}

\FloatBarrier
According to the Fig.\ref{Conf_L2_SVM}, Fig.\ref{Conf_L2_KNN} and Fig. \ref{Conf_L2_RF}, the Level 2 implementation of the Sri Lankan coin classification system demonstrates remarkable consistency across all three machine learning models - Random Forest, SVM, and KNN. Each model achieved identical performance metrics, showing perfect classification accuracy across all coin denominations. The confusion matrices reveal precise predictions for both old and new versions of the coins. The diagonal-only nature of all three confusion matrices indicates zero misclassifications, highlighting the robustness of the classification system in distinguishing between both denominations and versions (old/new) of Sri Lankan coins. This exceptional performance across all models suggests that the feature extraction and classification methodology effectively captures the distinctive characteristics of each coin category.
\FloatBarrier

\FloatBarrier
\FloatBarrier

\begin{figure}[!h]
	\centering
	\includegraphics[width=0.45\textwidth]{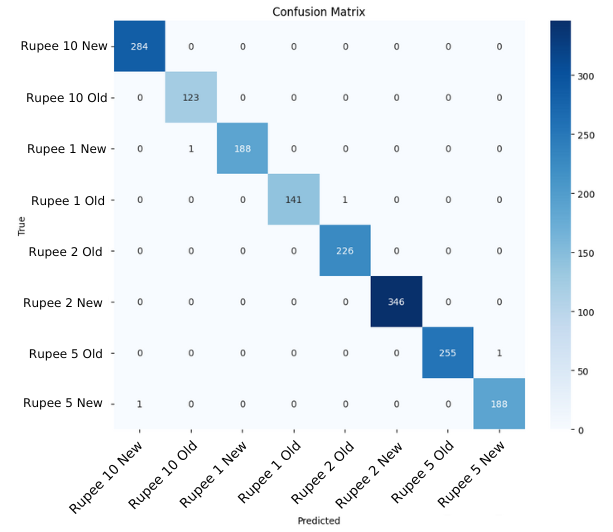}
	\caption{Confusion matrix of CNN}
	\label{Conf_L2_CNN}
\end{figure}

The custom CNN (Level 3) achieved exceptional classification accuracy with very minimal misclassifications across all denominations. The confusion matrix in Fig. \ref{Conf_L2_CNN} shows excellent classification performance with very few misclassifications. Most predictions fall on the diagonal, indicating high accuracy in distinguishing between old and new versions of each denomination. The total number of misclassifications is minimal, with only 4 instances of confusion across all classes. The results indicate that the custom CNN architecture successfully learned the distinctive features of both old and new versions of Sri Lankan coins, with nearly perfect classification accuracy.
\FloatBarrier
\FloatBarrier
\FloatBarrier

Comparing the performance across the different classification models, it is evident that SVM outperforms KNN and Random Forest in Level 1 implementation, showing consistent accuracy and strong performance across precision, recall, and F1-score. KNN also performs well but struggles slightly with certain denominations, while Random Forest exhibits the lowest performance, particularly for specific classes. In Level 2, all three models, including SVM, KNN, and Random Forest, achieve identical perfect classification results, suggesting that the dataset and feature extraction methodology are highly effective in distinguishing between coin types. Level 3 with the custom CNN model demonstrates the highest accuracy, surpassing all traditional classifiers with minimal misclassifications. The Sri Lankan coin dataset proves to be highly effective for training machine learning models. Traditional classifiers like SVM, KNN, and Random Forest provide a strong foundation, with SVM leading in traditional methods. However, deep learning techniques, as demonstrated by the CNN model, deliver the most precise results, reinforcing the power of deep learning in image classification tasks.

\section{Conclusion}
The Sri Lankan Coins Image Dataset represents a significant contribution to the field of computer vision and machine learning, particularly in the context of coin recognition and classification. This dataset, which is publicly accessible on Kaggle, comprises a diverse collection of images representing various denominations of Sri Lankan coins, categorized into both old and new variants. The dataset includes a total of 24,867 images, distributed across eight distinct classes, as detailed in Table \ref{tab:denomination_images}. With its comprehensive collection of images spanning multiple denominations and variants, the dataset provides a robust foundation for developing accurate and reliable coin recognition systems. The inclusion of both old and new coin variants ensures a broad representation, enabling models to capture the nuances of different designs and improve generalizability. This dataset holds immense potential for practical applications, particularly in enhancing the independence of visually impaired individuals during financial transactions. Future research can build on this foundation by exploring advanced techniques such as deep learning and expanding the dataset to include additional variations and environmental conditions. By doing so, the dataset can continue to drive innovation and contribute to the development of more effective and inclusive technological solutions.

\bibliographystyle{vancouver}
\bibliography{ref}
	
\end{document}